\documentclass[sigconf]{acmart}
\usepackage{CJKutf8}
\usepackage{multirow}
\usepackage[linesnumbered,ruled,vlined]{algorithm2e}
\usepackage{subcaption}

\AtBeginDocument{%
  \providecommand\BibTeX{{%
    \normalfont B\kern-0.5em{\scshape i\kern-0.25em b}\kern-0.8em\TeX}}}

\begin{document}

\title{SentPWNet: A Unified Sentence Pair Weighting Network for Task-specific Sentence Embedding}

\author{Li Zhang}
\affiliation{%
  \institution{Alibaba Group}
}
\email{lizhang.nus2010@gmail.com}

\author{Lingxiao Li}
\affiliation{%
Alibaba Group
}
\email{reagan.llx@alibaba-inc.COM}

\author{Han Wang}
\affiliation{%
Alibaba Group
}
\email{muwei.wh@alibaba-inc.COM}


\begin{abstract}
Pair-based metric learning has been widely adopted to learn sentence embedding in many NLP tasks such as semantic text similarity due to its efficiency in computation. Most existing works employed a sequence encoder model and utilized limited sentence pairs with a pair-based loss to learn discriminating sentence representation. However, it is known that the sentence representation can be biased when the sampled sentence pairs deviate from the true distribution of all sentence pairs. In this paper, our theoretical analysis shows that existing works severely suffered from a good pair sampling and instance weighting strategy. Instead of one time pair selection and learning on equal weighted pairs, we propose a unified locality weighting and learning framework to learn task-specific sentence embedding. Our model, SentPWNet, exploits the neighboring spatial distribution of each sentence as locality weight to indicate the informative level of sentence pair. Such weight is updated along with pair-loss optimization in each round, ensuring the model keep learning the most informative sentence pairs. Extensive experiments on four public available datasets and a self-collected place search benchmark with 1.4 million places clearly demonstrate that our model consistently outperforms existing sentence embedding methods with comparable efficiency. 
\end{abstract}





\begin{teaserfigure}
\includegraphics[width=\textwidth]{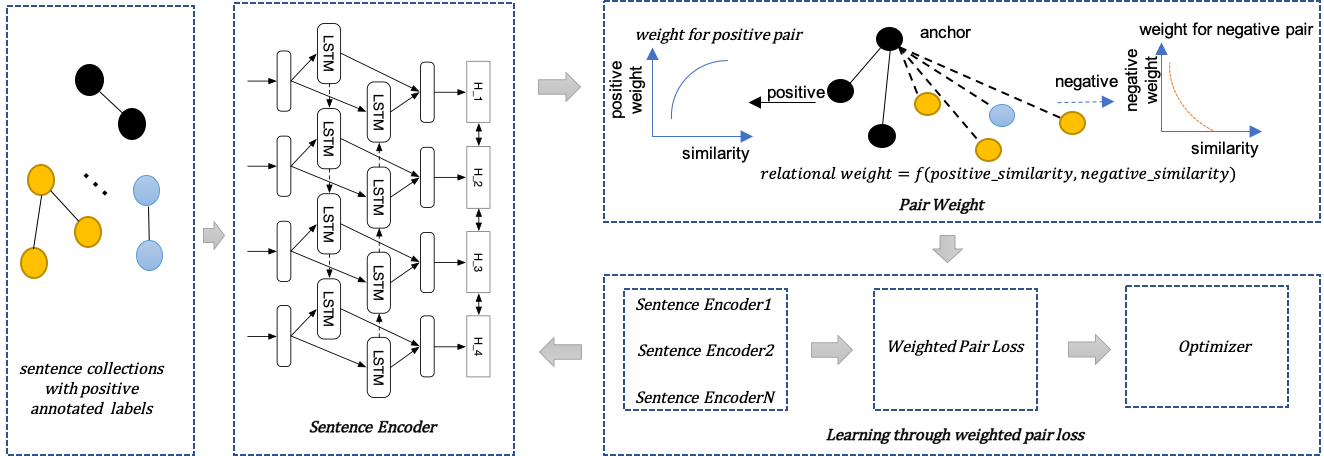}
\caption{An illustration of how SentPWNet learned task-specific sentence embedding. Instead of pre-sampling all the negative and positive pairs, SentPWNet had three key components: 1) a typical sequence encoder, such as BiLTSM. Our experimental results suggested that the choice of basic encoder did not significantly affect the performance of learned embedding. 2) locality weighting module. Instead of sampling negative pairs at beginning, SentPWNet employed a locality weight on-the-fly, where the weight was calculated based on its own similarity and its similarities of all their neighboring. Locality weighting technique was firstly introduced in computer vision~\cite{cai2007locality, wang2019multi}, but as far as we know, we are the first to integrate this technique into natural language processing and obtain quite promising results. 3) A pair-based loss optimizer to train the encoder based on the weighted pairs. Note that our framework is iterative, therefore SentPWNet can learn a better representation at each epoch until convergence. }
\label{fig:teaser}
\end{teaserfigure}

\maketitle

\section{Introduction}
Sentence embedding has attracted extensive attention for semantic text similarity since its wide usage in a broad range of NLP tasks, such as document organization and indexing, community question answering systems and large-scale information retrieval~\cite{aggarwal2012survey, ein2018learning}. For examples, in the platforms of cQA, such as Quora or Yahoo Answers, the community-driven nature of these platforms leads to a large amount of question duplication, therefore it is eager to have a way to identify similar paraphrase, which can reduce clutter and greatly improve the user experience. In general, there are two main model families to address semantic text similarity: 1) cross-encoder model that directly computed the similarity between a sentence pair without learning a sentence embedding explicitly. 2) a sentence embedding learner that tried to map a sentence into a real-value fixed-size representation, ensuring the similar sentences kept closer and dissimilar ones kept further. 

The main advantage of sentence embedding learner over cross-encoder models lies on the high efficiency in computation. thus it is in favor of many practical industry applications. For cross-encoder models~\cite{rao2019bridging, zhou2018multi, devlin2018bert, wang2017bilateral}, a sentence pair was required as input and the model can directly predict the target score, while no independent sentence embedding was computed. Given a collection of $~1,000,000$ sentences to find the most similar sentence, cross-encoder models had to compute the score for all one million pairs for each query. If BERT was employed, it would took around 20 hours on a Titan-X GPU for a single probe, which made this method completely infeasible despite its promising results. 

Sentence embedding remains an open yet challenging research problem. The extremely large number of sentences poses a great challenge to learn a discriminating sentence embedding. Assume vocabulary size is $K$ and the sentence length is $L$, the total number of entire sentence space is in the exponential magnitude of $K^L$. Specifically, recent works have explored many learning techniques with different training objectives to learn fixed-length sentence representations. Some works aims to extending the success of word embedding trained on large amounts of text in an unsupervised manner~\cite{mikolov2013distributed, pennington2014glove} and tries to exploit the sentence context to learn general-purpose sentence embedding that can directly be utilized in various downstream NLP tasks. These works, such as Skip-Thoughts~\cite{kiros2015skip}, FastSent~\cite{hill2016learning}, proposed to utilize an encoder-decoder architecture (seq2seq) to predict the contextual sentence from large corpus of articles. Nevertheless, these methods have severely suffered from insufficient training samples, thus the learned sentence embeddings were not performed well in many tasks. Some researchers~\cite{zhang2019bertscore, may2019measuring} started to input individual sentences into popular language models and derived fixed-size sentence embedding directly. The most commonly used approach was to perform an average pooling of whole output sequences or directly take the output of the first token (the [CLS] token). Unfortunately, this method has been shown the worse performance compared with averaging of simple word embedding in many works~\cite{reimers2019sentence, ein2018learning}. 

The rapid development of metric learning in computer vision~\cite{SchroffKP15,wang2019multi, HarwoodGCRD17, GeHDS18} stimulates a new direction for the researcher in NLP.  A few works ~\cite{conneau2017supervised, SubramanianTBP18, reimers2019sentence, ein2018learning} have attempted to combine typical sentence encoder with pair-based metric learning, such as contrastive loss~\cite{hadsell2006dimensionality} in Siamese network and triplet loss in Triplet Network, to learn sentence embedding and achieved quite promising results up to date. Nevertheless, these works have at least two limitations. Firstly, due to the difficulty of evaluate the dissimilarity of sentence pairs, existing works performed pair sampling randomly or utilized the distance of sentences in the article to perform pair sampling. Therefore, these works may neglect many informative pairs. Secondly, how the sampled sentence pairs affected the learned sentence embedding remain unknown, while researches in computer vision had shown that harder pairs were more informative and can drive better representation learning. 

In this paper, from the perspective of pair-loss optimization, our theoretical analyse verified the finding in computer vision and showed that the only two key components to learn sentence embedding were pair sampling and instance weighting. As far we know, this was the first work in NLP to clearly stating the limitation of existing works on sentence embedding that sentence pairs shall be selected and weighted with meticulous efforts. Our model, SentPWNet, tried to overcome these limitations by iteratively incorporating a locality preserving and weighted pair-loss optimizer. The framework of SentPWNet was shown in Figure~\ref{fig:teaser}. The novelty of SentPWNet were in two fold. Firstly, SentPWNet utilized the locality to measure the informative level of a sentence pair, similar to~\cite{GeHDS18,wang2019multi}. The locality weight was computed as the relative similarity between the similarity of each pair and the similarities to the others. Sentence pairs that have complex locality usually are hard to differentiate, thus they can get a larger weight and contribute more during the optimization, and vise versa. Secondly, the learning of SentPWNet was in an iterative learning manner, meaning that our model can benefit from the hardest pair sentences in each round until model convergence. To evaluate the performance, we conducted extensive experiments on three public benchmarks~(Quora, MRPC, Wikipedia Section and Wikipedia Title~\cite{ein2018learning}) on semantic text similarity and thematic relatedness tasks. The experiment results showed that SentPWNet is superior to existing sentence embedding with a marginal improvement on all tasks. Furthermore, with the popularity of local life services, such as Yelp and Meituan, we collected a new place search dataset with total 1.4 million point of interest (i.e. POI). As far as we know, this is the first POI dataset in million level that can provide a new benchmark for place search for the community, and we are going to make it public available in the near future. The experiment result on POI dataset indicated that our model was very effective in retrieval and consistently performed better than those baselines.

The paper is structured in the following way. Section~\ref{sec:rel} gives a review of related work. Section~\ref{sec:method} explains the locality-weighting in theory and illustrates the details of the proposed model, SentPWnet. Experimental results on five datasets, including Quora, MRPC, Wikipedia Sentence, Wikipedia Title and POI search dataset are demonstrated in Section~\ref{sec:experiments}. Finally, the conclusion and future work are presented in Section~\ref{sec: conclusion}.

\section{Related Works}
\label{sec:rel}
In this section, we firstly present a preliminary survey of existing sentence embedding and then introduce the usage of deep metric learning in NLP. 
\subsection{Sentence Embedding} 
Word embedding ~\cite{mikolov2013distributed, pennington2014glove} has driven a lot of success in  natural language processing (NLP), while it remains an open problem to extend this success to learning sentence representation. Various works have been explored unsupervised as well as supervised learning technique with different training objectives to learn fixed-length sentence representations. 

Since sentence consisted of multiple words, some works~\cite{le2014distributed, wieting2015towards,arora2016simple} regarded sentence embedding as a weighted summation of word embedding. The weight of each word can be pre-set or learned, such as averaging or smooth inverse frequency weight. This method provided an easy-yet-effective baseline for sentence embedding, but they did not exploit the syntax relations of words, which limited the discriminating ability of learned sentence embedding. Instead of averaging of word embedding, another unsupervised approach tried to utilize the sentence relation to train sentence embedding. Skip-Thoughts ~\cite{kiros2015skip} extended the skip-gram model on words to sentences level and trained an encoder-decoder architecture to predict the surrounding sentences. To address the efficiency of encoder-decoder model, FastSent~\cite{hill2016learning} replaced RNN architecture as embedding extractor into of a simple average of the word embedding, while Quick-Thought \cite{logeswaran2018efficient} changed encoder-decoder model to a classification task where the embedding of two sentences were classified whether they were adjacent or not. In summary, these works only utilized the sentence context in unsupervised manner, while neglecting the supervised signal for many sentence related tasks, thus they were limited by the poor performance of learned embedding.  Recently, some researchers~\cite{zhang2019bertscore,may2019measuring}  proposed to extend the success of language models into sentence embedding. They attempted to use the average of all output layer or the CLS token from BERT~\cite{devlin2018bert} as sentence embedding. Unfortunately, language models were trained to capture the essential relation between a sentence pair, thus it failed to get a satisfying performance for sentence embedding as proven in works~\cite{reimers2019sentence, ein2018learning}.

\subsection{Deep Metric Learning in NLP} 
As a fundamental machine learning task, metric learning has been widely applied in computer vision and natural language processing. The goal of metric learning aims to learn an embedding space, where the embedded vectors of similar ones are encouraged to be closer, while dissimilar ones are pushed apart from each other. In~\cite{cer2018universal}, a transformer network was trained for each sentence and a regression loss is concatenated to enforce the optimization of encoder. Another typical work was Sentence-Bert~\cite{reimers2019sentence}. This work proposed to utilize the embedding of BERT as the raw embedding and then fine-tuned it with Siamese and Triplet networks on NLI dataset. Other works ~\cite{neculoiu2016learning,ein2018learning,hoffer2015deep} were mostly similar to Sentence-Bert~\cite{reimers2019sentence}, except that they utilized stacked bi-LSTM or other basic encoder to replace BERT. It can be clearly seen that existing metric learning works in NLP focus on applying contrastive loss in Siamese network or triplet loss in Triplet network on sentence embedding learning, while lacking of in-depth study. In metric learning, it often generates highly redundant pairs, especially negative pairs, which are mostly uninformative. Inappropriate processing of these pairs can slow the convergence of model, thus yielded less discriminating embedding. Some works in computer vision~\cite{schroff2015facenet,harwood2017smart, kumar2010self, wang2019multi} have showed that better image embedding can be trained with more suitable pair mining and instance weighting. Motivated by this observation, we incorporate the characteristic of language, and present a novel pair weighting framework to learn sentence embedding. Our learned sentence embedding has been shown to be effective for each particular task.

\section{Method}
\label{sec:method}
In this section, we firstly illustrate the importance of pair sampling and instance weighting in theory, which can clearly indicate the limitation of existing works on sentence embedding. We then introduce the details of our proposed locality weight framework, SentPWNet.

\subsection{Pair Sampling and Instance Weighting}
Similar to recent works in computer vision~\cite{schroff2015facenet, wang2019multi}, we provide a deep analysis of how metric learning learns the embedding from the perspective of loss optimization.


Given a batch of sentences and corresponding labels $(X \in R^{m*l}, Y \in R^m)$, $m$ for the number of training samples and $l$ for the length of sentence. Then a sentence $x_i$ is encoded as a vector $v_i$ by a neural network encoder parameterized by $\theta$. Note there is not too much restriction of such encoder.  The similarity of two samples is $S_{ij}=<v_i, v_j>$. Given a batch of sentences, it can generate an  $m \times m$ similarity matrix $S$,  whose element at $(i, j)$ is $S_{ij}$.

Given an arbitrary pair-based loss $\mathcal{L}$, it can be formulated as a function in terms of $S$ and $y: \mathcal{L}(S,y)$. The derivative with respect to model parameters $\theta$ at the t-th iteration can be calculated as:

\begin{equation}
\label{equ:1}
\frac{\partial{\mathcal{L}(S,y)}}{\partial{\theta}}\bigg|_t = \frac{\partial{\mathcal{L}(S,y)}}{\partial{S}}\bigg|_t \frac{\partial{S}}{\partial{\theta}}\bigg|_t
= \sum_{i=1}^m \sum_{j=1}^m \frac{\partial{\mathcal{L}(S,y)}}{\partial{S_{ij}}}\bigg|_t \frac{\partial{S_{ij}}}{\partial{\theta}}\bigg|_t
\end{equation}

\noindent while equation \ref{equ:1} has the same gradient with respect to $\theta$ as equation \ref{equ:2} at the t-th iteration, which is formulated as below:
\begin{equation}
\label{equ:2}
\mathcal{F}(S, y) = \sum_{i=1}^m \sum_{j=1}^m \frac{\partial{\mathcal{L}(S,y)}}{\partial{S_{ij}}}\bigg|_t S_{ij}
\end{equation}

$\frac{\partial{\mathcal{L}(S,y)}}{\partial{S_{ij}}}\bigg|_t$ is a constant scalar as it is not involved in the gradient of equation \ref{equ:2} with respect to $\theta$. As we show later, $\frac{\partial{\mathcal{L}(S,y)}}{\partial{S_{ij}}}\bigg|_t$ varies from different pair-loss definitions. 

For a pair-based pair loss $\mathcal{L}$, we can assume that $\frac{\partial{\mathcal{L}(S,y)}}{\partial{S_{ij}}}\bigg|_t \geq 0$ for a negative pair and $\frac{\partial{\mathcal{L}(S,y)}}{\partial{S_{ij}}}\bigg|_t \leq 0$  for a positive pair. Thus, equation \ref{equ:2} can be transformed into following pair weighting formulation:
\begin{equation}
\label{equ:3}
\begin{aligned}
\mathcal{F} &= \sum_{i=1}^m (\sum_{y_j\not= y_i}^m \frac{\partial{\mathcal{L}(S,y)}}{\partial{S_{ij}}} \bigg|_t S_{ij} + \sum_{y_j = y_i}^m \frac{\partial{\mathcal{L}(S,y)}}{\partial{S_{ij}}} \bigg|_t S_{ij})\\ &= \sum_{i=1}^m (\sum_{y_j\not= y_i}^m w_{ij}{S_{ij}} - \sum_{y_j = y_i}^m w_{ij}{S_{ij}})
\end{aligned}
\end{equation}

where the weight for any pair {$x_i, x_j$} is $w_{ij} = \bigg|\frac{\partial{\mathcal{L}(S,y)}}{\partial{S_{ij}}}\big|_t\bigg|$. In such way, the loss of any pair-based model can be formulated as weighting for pair-wise similarities.

\subsection{Revisiting Existing Metric Learning on Sentence Embedding}
We revisit the existing related works on using metric learning. In general, they~\cite{neculoiu2016learning,ein2018learning,hoffer2015deep,reimers2019sentence} are using two popular pair-based loss functions:  soft margin triplet loss~\cite{hermans2017defense} in Triplet network and contrastive loss~\cite{hadsell2006dimensionality} in Siamese Network.

\noindent\textbf{Soft margin triplet loss.} Soft margin triplet loss was first proposed in~\cite{ein2018learning} to learn sentence embedding on thematic similarity. It replaces the traditional hinge function by a smooth approximation using the softplus function:
\begin{equation}
L_{triplet}=log(1+e^{S_{an}-S_{ap}})
\end{equation}
where $S_{an}$ and $S_{ap}$ denote the similarity of a negative pair {$x_a$, $x_n$}, and a positive pair {$x_a$, $x_p$}, with respect to an anchor sample $x_a$. By computing partial derivative with respect to $S_{ij}$ in equation~\ref{equ:2}, we can find that triplet loss assigns all sampled pairs equal weights, where
\begin{equation}
w_{ij}=1,
\end{equation}
for all sampled pairs. For those pairs that are threw away, it does not make any contribution to the embedding learning, thus can be viewed as zero weight, $w_{ij}=0$.

\noindent\textbf{Contrastive loss.} This loss has been widely in Siamese network. It aims to take positive pairs closer and push negative pairs apart from each other. It is defined as
\begin{equation}
\mathcal{L}_{contrastive} = (1-\mathcal{I}_{ij})[S_{ij}-\lambda]_+-\mathcal{I}_{ij}S_{ij}
\end{equation}
where $\mathcal{I}_{ij}$=1 indicates a positive pair, and 0 for a negative one. By computing weights of selected pairs, all selected positive pairs and hard negative pairs ($S_{ij}>\lambda$) are weighted equally, where
\begin{equation}
S_{ij}=1
\end{equation}
For those pairs that are filtered out, the weights for them can be regarded as zero.

\subsection{SentPWNet}
We address two obvious limitations in existing works: 1) The sampling is fixed, thus many informative pairs are discarded. 2) Pair instances may have different locality structures, so it is unsuitable to treat them equally.  
To address these two problems, we propose a new model, SentPWNet, that adopts a locality weighting technique for each sentence pair. This weighting is not new in computer vision~\cite{wang2019multi}, but as far as we know, we are the first to apply it on natural language processing and push the frontier of metric learning research in NLP. Actually, the locality weighting is quit suitable in NLP, because the meaning of sentence greatly relies on its contextual sentences. The locality weight is formulated as:
\begin{equation}
\label{equ:ms}
\begin{aligned}
\mathcal{L}_{MS} = \frac1m \sum_{i=1}^m \{\frac{1}{\alpha}log[1+\sum_{k \in \mathcal{P}_i}e^{-\alpha(S_{ik}-\lambda)}]\} \\
+ \frac1\beta log[1+\sum_{k \in \mathcal{N}_i}e^{\beta(S_{ik}-\lambda)}]]
\end{aligned}
\end{equation}
where $\mathcal{P}_i$ means positive samples and $\mathcal{N}_i$ means negative samples given an anchor $i$, and $\alpha$, $\beta$, $\lambda$ are hyper-parameters.

\noindent\textbf{Simultaneous Pair weighting and Sampling.} SentPWNet does not have an independent pair mining stage. Instead, it gives weight for each sentence pair. Based on equation \ref{equ:ms} , given any selected negative pair $\{x_i, x_j\} \in \mathcal{N}_i$ , its weight $w_{ij}^-$ is computed as:

\begin{equation}
w_{ij}^- =  \frac1{e^{\beta(\lambda-S_{ij})}+\sum_{k \in \mathcal{N}_i}e^{\beta(S_{ik}-S_{ij})}}
\end{equation}

\noindent and the weight $w_{ij}^+$  of $\{x_i, x_j\} \in \mathcal{P}_i$ is computed as :

\begin{equation}
w_{ij}^- =  \frac1{e^{-\alpha(\lambda-S_{ij})}+\sum_{k \in \mathcal{P}_i}e^{-\alpha(S_{ik}-S_{ij})}}
\end{equation}

From the equations above, the weight of a negative pair is computed jointly from its own similarity by $e^{\beta(\lambda-S_{ij})}$ and relative similarity by $e^{\beta(S_{ik}-S_{ij})}$, and the positive pair is similar. 

Given an anchor, $x_i$, we find that the pair whose locality has been already well preserved in current sentence embedding space can directly be assigned a negligible weight. This can save up to $50\%$ computation based on our experiments. For simplicity, we regard the locality of a pair has not been fully preserved, if and only if it satisfy two conditions,
1) for a negative pair ${x_i, x_j}$ if :
\begin{equation}
S_{ij}^- >  \underset{y_k=y_i}{min} S_{ik} - \epsilon
\end{equation}
2) for a positive pair:
\begin{equation}
S_{ij}^+ <  \underset{y_k \not= y_i}{max} S_{ik} + \epsilon
\end{equation}
where $\epsilon$ is a hyper parameter. We usually set $\epsilon$ to be 0.1.

\section{Experiments}
\label{sec:experiments}
In this section, we first introduce the experiment settings. Then we demonstrate the properties of SentPWNet through three public benchmarks on typical semantic text similarity and thematic relatedness tasks. Moreover, we propose a self-collected place search benchmark as entity matching task, and evaluate SentPWNet on it. Finally, we show the visualizations of our experimental results to illustrate the model performance.

\subsection{Experiment Settings}
\label{section1}

For model architecture, we respectively use typical CNN, LSTM (same as ~\cite{wang2017bilateral}) and BiLSTM (same as ~\cite{ein2018learning}) networks as encoders to embed sentences. All the input sentences are split and initialized in the word representation layer with the 300-dimensional GloVe word vectors pre-trained from the 840B Common Crawl corpus~\cite{pennington2014glove}, and the out-of-vocabulary words' embeddings are initialized randomly. For simplicity, we use pairwise cosine similarities in the embedding space to evaluate all the sentence embedding methods.

SentPWNet adopts a locality weighting technique to iteratively learn all informative pairs and add more weight to more informative pairs. To demonstrate the importance of the properties, we conduct two ablation studies. 1) To investigate the impact of informative pairs. 2) To verify the weighting mechanism. Moreover, later in this section we will show the speed advantage of SentPWNet compared with cross-encoder models. We further compare the performance of our method with the state-of-the-art techniques on semantic text similarity and thematic relatedness tasks.



\subsection{Sentence Embedding Tasks}
\label{section3}

To demonstrate that our model consistently outperforms existing sentence embedding methods, we further compare SentPWNet with state-of-the-art on classical semantic text similarity and thematic relatedness tasks.

\subsubsection{Semantic Text Similarity}
\label{section1}

\begin{table*}[!h]
\centering
\caption{Examples of QQP dataset and MRPC. label=1 or 0 indicates the sentence pair is semantic similar or not.}
\label{tab:quora-mrpc-examples}
\begin{tabular}{llll}
\toprule
\textbf{ }&\textbf{sentence 1}&\textbf{sentence 2}& \textbf{label} \\
\midrule
\multirow{2}{*}{QQP} &  How can I avoid sleeping in a boring class ?       & How do I not sleep in a boring class ?                   & 1 \\
                       &  How much does it cost to fix a scratched bumper ?  & How do i prevent scratching the front bumper of my car ? & 0 \\
\midrule
\multirow{2}{*}{MSRP}  &  I'm never going to forget this day.                & I am never going to forget this throughout my life.      & 1 \\
                       &  Looking to buy the latest Harry Potter?            & Harry Potter's latest wizard trick?                      & 0 \\
\bottomrule
\end{tabular}
\end{table*}

Semantic similarity task deal with determining whether two sentences are semantically consistent, such as answer sentence selection and paraphrase identification. Specifically, for paraphrase identification, $x_i$ and $x_j$ are two sentences, $Y \in \{0, 1\}$, where $y = 1$ indicates that $x_i$ and $x_j$ are paraphrase of each other, and $y = 0$ otherwise.

\noindent\textbf{Datasets.} We evaluate our model on quora question pair (QQP) paraphrase dataset and Microsoft Research Paraphrase Corpus (MRPC). QQP dataset contains over 400K question sentence pairs from the questions on quora website, and each pair is annotated with a binary value (1 or 0) indicating whether the two questions are paraphrase of each other (positive or negative). And the ratio of positive and negative sentence pairs is $1: 1$. The data split of QQP is same as \cite{wang2017bilateral}\footnote{This split is available at https://zhiguowang.github.io.}, with 10K question pairs each for development and testing, the remaining instances are used as training set. MRPC contains 5800 sentences pairs extracted from news sources on the web, also along with binary values indicating whether each pair captures a semantic equivalence relationship. We show some examples of QQP dataset and MRPC in Table \ref{tab:quora-mrpc-examples}.




During the test stage, we use a threshold search method to find appropriate similarity threshold and then use it to distinguish whether the input pair is identical. Specifically, we use a threshold set to search a threshold value, that is, a sentence pair whose similarity is higher than the threshold is regarded as a positive pair, otherwise as negative pairs:

\begin{algorithm}
\SetAlgoLined
\KwIn{training set $Q$, testing set $P$, iterations $N$, threshold set $T$, similarity threshold space $S\in(0, 1)$}
\KwOut{accuracy and threshold set of each iteration on $P$}
\For{$i=0$ \KwTo $N$}{
  train model on $Q$ ...\;
  \tcc{testing stage}
  initialize the accuracy of iteration $i$ as $Acc_i=0$\;
  \While{$S$ has next}{
    get $S_k=S.next$\;
    use $S_k$ as threshold to compute the $accuracy$ on $P$\;
    \eIf{accuracy>$Acc_i$}{
      $Acc_i=accuracy, Threshold_i=S_k$\;
    }{continue\;}
  }
}
\caption{Threshold search method}
\end{algorithm}


For QQP dataset, we use the development set as the threshold set, and for MRPC, we use the whole training set. 

\noindent\textbf{Evaluation and Analysis.} We compare our method with the advanced pre-trained methods and two typical deep metric learning (DML) contrast methods. The pre-trained methods include BERT~\cite{devlin2018bert}, Skip-Thoughts~\cite{logeswaran2018efficient} with the bi-skip model, InferSent~\cite{conneau2017supervised} trained with Glove and Sentence-BERT~\cite{reimers2019sentence}, and the aim of assessing them is to examine how well the state-of-the-art general-purpose methods perform on the specific task. The DML methods are implemented under the well-known Siamese framework with contrastive loss and triplet framework with triplet loss, both of them use the same encoders of SentPWNet.

For Triplet network and SentPWNet, we regard all the sentences of a batch which don't share the identical intent with anchor sentence as negative samples. We use hard sample mining~\cite{SchroffKP15} to construct training batches for Triplet network, and SentPWNet do not need to collect informative sentence pairs in advance.

Table \ref{tab:QQP+MRPC} shows the performance of the baseline methods and SentPWNet. First, it's obvious that general-purpose methods could not achieve satisfying performance compared with DML methods, which reveals that general-purpose embedding fail to generalize to paraphrase identification task and it is necessary to derive task-specific sentence embeddings. Second, no matter which encoder is used, SentPWNet increases the accuracy and f1-score by average 1\% on QQP dataset and MSRP compared with other DML methods. It tells us that the locality weighting scheme is more effective than equal weighting ways. Furthermore, despite the fact that the independent pair mining stage such as hard sample mining do sample many informative pairs, our method can still mine more informative pairs during a single weighting stage to achieve better performance. In conclusion, our method is effective for paraphrase identification task.


\begin{table}[!h]
\centering
  \caption{Evaluation on QQP dataset and MRPC.}
  \label{tab:QQP+MRPC}
  \begin{tabular}{l|cc|cc}
\toprule
~ & \multicolumn{2}{c}{QQP dataset} & \multicolumn{2}{c}{MRPC}\\
\hline
\textbf{Model} & Acc & F1 & Acc & F1\\
\hline
    Avg.BERT embeddings  & 0.698 & 0.723 & 0.698 & 0.783 \\
    BERT cls-vector & 0.676 & 0.693 & 0.634 & 0.726 \\
    Skip-Thoughts & 0.678 & 0.705 & 0.640 & 0.775 \\
    InferSent & 0.692 & 0.732 & 0.695 & 0.790 \\
    Sentence-BERT & 0.730 & 0.739 & 0.731 & 0.812 \\
\hline
    BiLSTM + Siamese network & 0.839 & 0.837 & 0.725 & 0.814 \\
    BiLSTM + Triplet network & 0.826 & 0.828 & 0.734 & 0.805 \\
    BiLSTM + SentPWNet & \textbf{0.850} & \textbf{0.854} & \textbf{0.743} & \textbf{0.819} \\
\hline
    CNN + Siamese network & 0.837 & 0.838 & 0.722 & 0.810 \\
    CNN + Triplet network & 0.811 & 0.818 & 0.725 & 0.808 \\
    CNN + SentPWNet & \textbf{0.858} & \textbf{0.863} & \textbf{0.736} & \textbf{0.825} \\
\hline
    LSTM + Siamese network & 0.834 & 0.836 & 0.721 & 0.812 \\
    LSTM + Triplet network & 0.806 & 0.809 & 0.735 & 0.815 \\
    LSTM + SentPWNet & \textbf{0.843} & \textbf{0.851} & \textbf{0.740} & \textbf{0.821} \\
\bottomrule
\end{tabular}
\end{table}

\subsubsection{Thematic relatedness}
\label{section2}

Thematic relatedness task is important for various applications, such as multi-document summarization and multi-document summarization. It deal with determining whether two sentences are thematically related. For instance, a sentence is thematically closer to sentences within its section than to sentences from other sections. We conduct experiments on two thematic relatedness datasets: Wikipedia section sentence triplets and Wikipedia section title triplets~\cite{ein2018learning}.


\noindent\textbf{Datasets.} ~\cite{ein2018learning} used Wikipedia to create two thematically fine-grained datasets for thematic relatedness task. Specifically, Wikipedia articles are divided into sections focusing on different themes. They collected anchor and positive examples from same sections, and negative sentence from the previous or next section in the same article, aiming to obtain more difficult and informative negative samples. And the construction of Wikipedia title triplets dataset is similar, where in each triplet the first sentence in the section is paired with the section title, as well as with the title of the previous/next sections (if exists), where the former pair is assumed to have greater thematic similarity. The sentence dataset has $1.8M$ training triplets and 222K testing triplets, and the title dataset has $1.38M$ training triplets and $172K$ testing triplets. Moreover, When we train Siamese network, we split a triplet into a positive pair and a negative pair.

\begin{table}[!h]
  \caption{Evaluation on Wikipedia section sentence and Wikipedia section title triplets}
  \label{tab:sentence+title}
  \begin{tabular}{l|c|c}
\toprule
~ & \ sentence & title \\
\hline
\textbf{Model} & Acc & Acc \\
\hline
    Avg.BERT embeddings & 0.687 & 0.569 \\
    BERT cls-vector & 0.677 & 0.572 \\
    Skip-Thoughts & 0.577 & 0.538 \\
    InferSent & 0.627 & 0.539 \\
    Sentence-BERT & 0.645 & 0.636 \\
\hline
    BiLSTM + Siamese network & 0.729 & 0.769\\
    BiLSTM + Triplet network & 0.731 & 0.770\\
    BiLSTM + SentPWNet & \textbf{0.733} & \textbf{0.785} \\
\bottomrule
\end{tabular}
\end{table}




\noindent\textbf{Evaluation and Analysis.} We augment negative pairs base on the original dataset. For the title triplets dataset, we treat all sentences that belong to different sections as negative samples. And for the sentence triplets dataset, we find that one sentence might appear in different sections, which make it incorrect to treat sentences from other sections as negative samples. In case of introducing noise, we only use the original negative samples provided by the sentence triplet dataset. In this way, there are at most two negative samples for each anchor.

We use accuracy as the evaluation measure, the accurate prediction is decided by whether the similarity between positive and anchor is greater than the similarity between negative and anchor. The results are shown in Table \ref{tab:sentence+title}. Note that in both two datasets, all the DML methods work much better than the general-purpose embedding methods by 10\% at least. For the sentence triplet dataset, because there are not enough negative samples, our model could not mine and weight informative pairs properly. So it's reasonable that SentPWNet achieves almost same performance as Siamese network and Triplet Network. For the title triplet dataset, there are thousands of pairs for one anchor, SentPWNet can obtain more informative pairs during optimizing stage and weight them appropriately, therefore our model outperforms Siamese network and Triplet network by 2\% above. The results demonstrate the effectiveness of locality weighting schema of SentPWNet. In conclusion, our model is also effective for thematic relatedness task.

\subsection{POI Entity Matching Dataset}
\label{section3}

With the popularity of local life services, such as outside catering and navigation service, we collected a new place search dataset with total 1.4 million point of interest (i.e. POI). Given a POI query, the task is to search the most similiar POI from the gallery. And the search process can be divided into two steps: 1) recall the candidates from total dataset; 2) match the duplicate POI from the candidates (if it exists). Here we extract the second step as a POI entity matching task, namely determining a pair of POIs are duplicate or not.

\begin{table}
  \caption{The category distribution of POI-EM-CHN dataset }
  \label{tab:sentence+title}
  \begin{tabular}{l|c|c}
\toprule
~ & \ sentence & title \\
\hline
\textbf{Model} & Acc & Acc \\
\hline
    Avg.BERT embeddings & 0.687 & 0.569 \\
    BERT cls-vector & 0.677 & 0.572 \\
    Skip-Thoughts & 0.577 & 0.538 \\
    InferSent & 0.627 & 0.539 \\
    Sentence-BERT & 0.645 & 0.636 \\
\hline
    BiLSTM + Siamese network & 0.729 & 0.769\\
    BiLSTM + Triplet network & 0.731 & 0.770\\
    BiLSTM + SentPWNet & \textbf{0.733} & \textbf{0.785} \\
\bottomrule
\end{tabular}
\end{table}

\noindent\textbf{Datasets.} The POI entity matching dataset is collected driven from place search scenario in map services. Each POI is defined by five key attributes: category, name, address, latitude and longitude, as shown in Table \ref{tab:poi_sample}. 

In practice we find that name attribute is very important for identification of POI.

\begin{itemize}
\item{}Utilize Geohash to process latitude and longitude into a hash code of length 19.
\item{}Divide name, address, and code into single characters.
\item{}Following the works in~\cite{TrisedyaQZ19}, we concatenate the segmented address, name, and hash code as a sentence.
\end{itemize}

The train set has a total of 1.41M examples, which belong to 537K classes. The test set has 12K examples, which belong to 5.8K classes. The entity matching task is to rank a list of candidate entity sentences based on their similarities to the anchor entity, and the performance of model is measured by Hit@n.

\begin{CJK*}{UTF8}{gkai}
\begin{table}[!h]
  \caption{An example of POI dataset}
  \label{tab:poi_sample}
  \begin{tabular}{l|p{6cm}}
\toprule
\textbf{Category} & 1000006 \\
\hline
\textbf{Name} & 致青春奶茶店 \\
\hline
\textbf{Address} & 武汉科技大学城市学院食代铭美食城c09 \\
\hline
\textbf{Latitude} & 30.5890440 \\
\hline
\textbf{Longitude} & 114.4297680 \\
\hline
\textbf{Hashcode} & 3760125996951404544 \\
\bottomrule
\end{tabular}
\end{table}
\end{CJK*}

\begin{table}[!h]
  \caption{Evaluation on POI retrieval dataset}
  \label{tab:poi}
  \begin{tabular}{lccc}
\toprule
\textbf{Model} & \textbf{Hit@1} & \textbf{Hit@3} & \textbf{Hit@10}\\
\hline
    Avg.BERT embeddings & 0.430 & 0.560 & 0.640\\
    BERT cls-vector & 0.252 & 0.350 & 0.435\\
\hline
    BiLSTM + Siamese network & 0.808 & 0.959 & 0.984\\
    BiLSTM + Triplet network & 0.799 & 0.954 & 0.981\\
    BiLSTM + SentPWNet & \textbf{0.819} & \textbf{0.971} & \textbf{0.988}\\
\hline
    CNN + Siamese network & 0.805 & 0.959 & 0.982\\
    CNN + Triplet network & 0.800 & 0.955 & 0.982\\
    CNN + SentPWNet & \textbf{0.810} & \textbf{0.965} & \textbf{0.985}\\
\hline
    LSTM + Siamese network & 0.811 & 0.963 & 0.984\\
    LSTM + Triplet network & 0.802 & 0.965 & 0.980\\
    LSTM + SentPWNet & \textbf{0.815} & \textbf{0.967} & \textbf{0.985}\\
\bottomrule
\end{tabular}
\end{table}

\noindent\textbf{Evaluation and Analysis.} For any anchor sentence, we regard all sentences having different label with anchor as negative samples. We use hard sample mining to derive pairs and triplets for Siamese network and Triplet network.

Table \ref{tab:poi} shows the performance of all methods. general-purpose sentence embedding models\footnote{POI dataset used here is composed of Chinese. We don't evaluate other general-purpose embedding methods because of its lack of support for Chinese corpora} still fail to achieve satisfying performance, and we can see that our method outperforms other methods by at least 0.4\% , which proves that our method is also effective for entity matching task.

\subsection{2D Visualization Of Features}
\label{section4}
\begin{figure*}[h!]
  \centering
  \begin{subfigure}[b]{0.33\linewidth}
    \includegraphics[width=\linewidth]{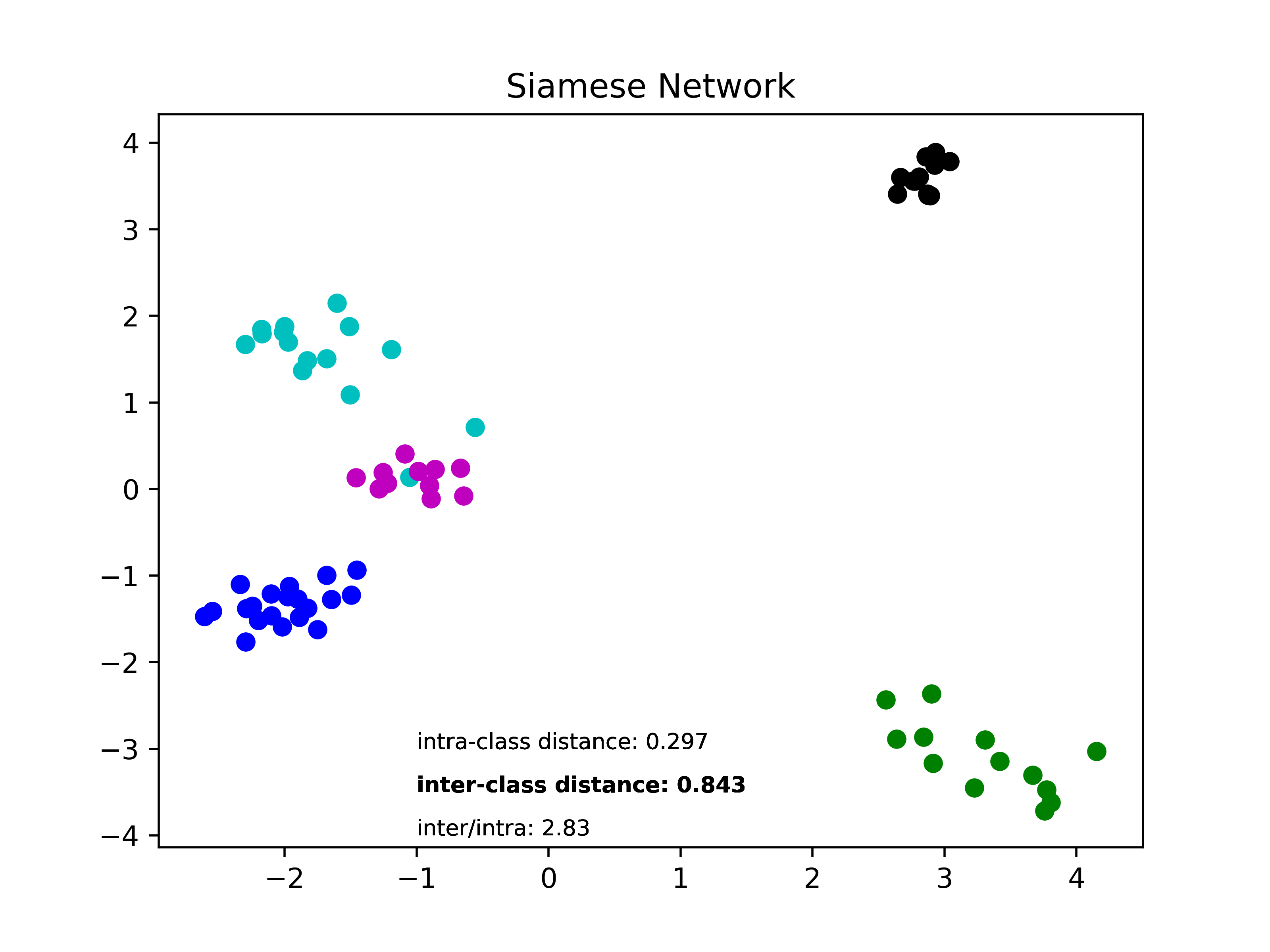}
    \caption{Siamese Network}
  \end{subfigure}
  \begin{subfigure}[b]{0.33\linewidth}
    \includegraphics[width=\linewidth]{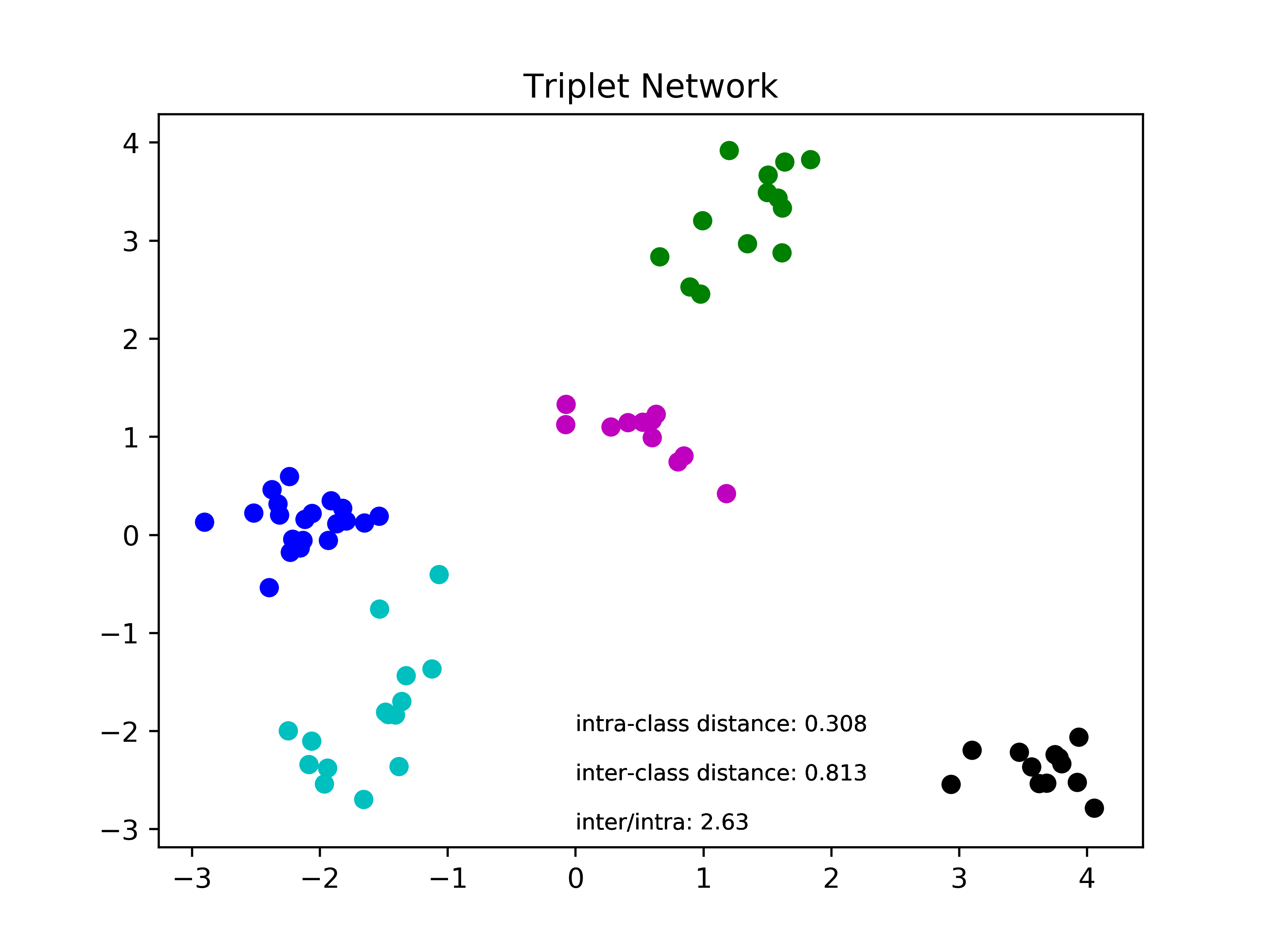}
    \caption{Triplet Network}
  \end{subfigure}
  \begin{subfigure}[b]{0.33\linewidth}
    \includegraphics[width=\linewidth]{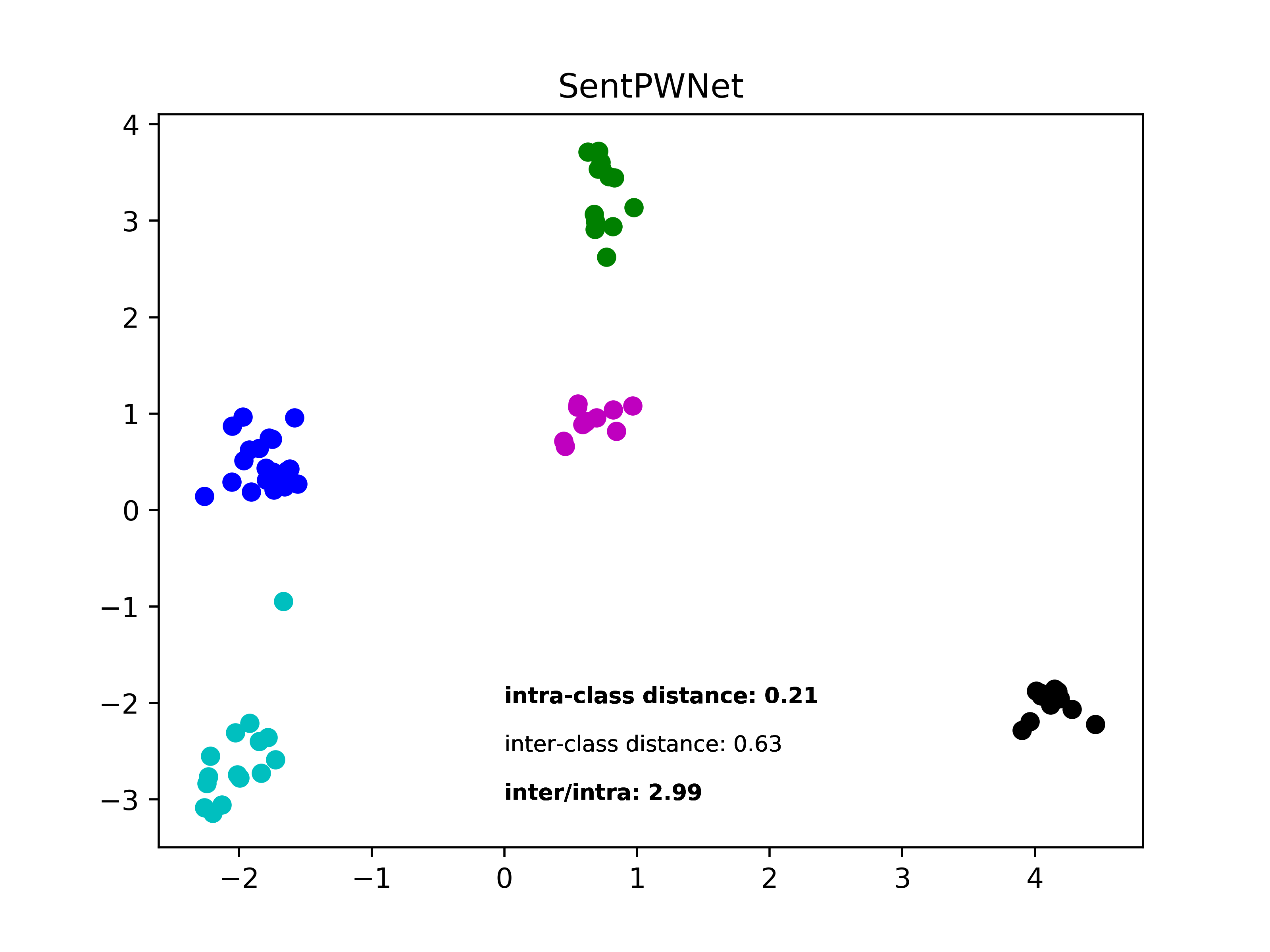}
    \caption{SentPWNet}
  \end{subfigure}
  \caption{Euclidean representation of embedded test data produced by BiLSTM encoder trained with different methods, Siamese Network, Triplet Network, SentPWNet. The projection was performed by T-SNE.}
  \label{fig:fig}
\end{figure*}

In order to intuitively prove that the sentence embeddings from SenPWNet are able to learn similarity relationship, we visualize the embeddings as 2D pictures. Concretely, we randomly select some sentences belonging to 5 classes in POI retrieval \textbf{test} set (table~\ref{tab:test}) and embed them with BiLSTM network which have been trained for 100 epochs under three different methods separately. Then we use PCA to project their embeddings into 2D euclidean space which can be easily visualized (Figures ~\ref{fig:fig}). We take the mean-pooling of all the sentence embeddings in each class as the center of the class. Moreover, we take the average of the distance between the sentence embedding and its class center as intra-class distance, and we take the average of the distance between the class center and the other closest class centers as inter-class distance. In this way, the larger the ratio of inter/intra, the better the clustering effect of the sample points in the vector space.

As shown in \ref{fig:fig}, sentence pairs' similarities can be easily measured by the distance between their embeddings, and a simple linear classifier might reach high classification accuracy. The inter/intra of SentPWNet is 2.99, which is better than Siamese Network and Triplet Network. It demonstrates that our model learned well to capture similarity relationship, ensuring the similar sentences kept closer and dissimilar ones kept further.


\begin{CJK*}{UTF8}{gkai}
\begin{table*}[!h]
  \caption{Examples of the selected classes, we connect the address, name and hashcode with <PAD> token.}
  \label{tab:test}
  \begin{tabular}{l|l}
\toprule
\textbf{Label} & \textbf{Sentence} \\
\hline
    1024033 & 南新街北段与台阶路交叉口西南150米 <PAD> 利群饸饹店 <PAD> 3919372159020957696\\
    1024033 & 南新街北段与台阶路交叉口西北50米 <PAD> 小杨饸饹 <PAD> 3919372159557828608\\
\hline
    107999 & 城南道38-40号地铺 <PAD> 柠檬泰国餐厅 <PAD> 3748128673543225344\\
    107999 & 九龙城城南道48号地下 <PAD> 泰象馆 <PAD> 3748128669416030208\\
\hline
    493600 & 中区明洞7街21(近乐天百货) <PAD> 烤肉明洞(本店) <PAD> 3854134548551958528\\
    493600 & 首尔市中区退溪路105 <PAD> isaac吐司明洞店 <PAD> 3854134551471194112\\
\hline
    612047 & 东京都中央区银座5丁目8-20银座コアb1f,〒1040061 <PAD> 小豆岛大仪银座店 <PAD> 6924438286450556928\\
    612047 & 日本〒106-0041东京都中央区银座6-5-15能楽堂ビル3f <PAD> 百菜百味银座店 <PAD> 6924438283631984640\\
\hline
    618086 & 东京都港区六本木4-9-8优座ビルb1 <PAD> 土风炉六本木店 <PAD> 6924437782026780672\\
    618086 & 东京都港区六本木3‐11‐2rosetokyo1f <PAD> 元祖六本木店 <PAD> 6924437782261661696\\
\bottomrule
\end{tabular}
\end{table*}
\end{CJK*}

\section{Conclusion and Future work}
\label{sec: conclusion}
In this paper, we are pushing the frontier of metric learning in NLP. Our theoretical analysis from the perspective of loss optimization provides a novel insight on the usage of pair-based that clearly indicates the importance of pair mining and instance weighting to learn sentence embedding. These two parts have been severely overlooked by most existing works. Our model, SentPWNet, incorporates the locality weighting schema and turns the conventional works with two isolated stages, sampling and learning, into a unified locality weighting and pair-based optimizing framework in an iterative manner. The experimental results clearly show the effectiveness of our model. Moreover, our self-collected POI dataset can provide the community a testbed for place retrieval task. 

For future work, there are many works to be exploited. Despite our locality weighting scheme gives relatively good performance, it is still unknown whether it is the optimal way. Another interesting direction is the interpretability of the learned representation. Our model is in supervised manner and relies on human annotated training samples to a large extent. Therefore, how to explain the semantic meaning of learned representation still requires a lot of future efforts. 
\begin{acks}
To Robert, for the bagels and explaining CMYK and color spaces.
\end{acks}

\bibliographystyle{ACM-Reference-Format}
\bibliography{sample-base}

\appendix

\end{document}